\pdfoutput=1

\documentclass[11pt]{article}

\usepackage{naacl2021}

\usepackage{times}
\usepackage{latexsym}

\usepackage[T1]{fontenc}

\usepackage[utf8]{inputenc}

\usepackage{microtype}

\usepackage{latexsym}
\usepackage{cleveref}
\usepackage{multirow}
\usepackage{adjustbox}
\usepackage{booktabs}
\usepackage{graphicx}
\usepackage{url}
\usepackage{array}
\usepackage{subcaption}
\newcolumntype{H}{>{\setbox0=\hbox\bgroup}c<{\egroup}@{}}
\usepackage{pdfpages}
\usepackage{pgffor}

\usepackage[main=english,russian]{babel}  

\usepackage{changepage}

\newlength{\offsetpage}
{\end{adjustwidth}}

\newcommand{\comment}[1]{}

\makeatletter
\newcommand\footnoteref[1]{\protected@xdef\@thefnmark{\ref{#1}}\@footnotemark}
\makeatother

\title{mT5: A Massively Multilingual Pre-trained Text-to-Text Transformer}

\author{Linting Xue\thanks{\hspace{0.5em}Equal Contribution. Please direct correspondence to \texttt{lintingx@google.com}, \texttt{nconstant@google.com}, \texttt{adarob@google.com}, and \texttt{craffel@google.com}} \quad {\bf Noah Constant}\footnotemark[1] \quad {\bf Adam Roberts\footnotemark[1]} \\ {\bf Mihir Kale} \quad {\bf Rami Al-Rfou} \quad {\bf Aditya Siddhant} \quad {\bf Aditya Barua} \quad {\bf
Colin Raffel} \\
Google Research}

\date{}

\begin{document}
\twocolumn
\maketitle

\begin{abstract}
The recent ``Text-to-Text Transfer Transformer'' (T5) leveraged a unified text-to-text format and scale to attain state-of-the-art results on a wide variety of English-language NLP tasks.
In this paper, we introduce mT5, a multilingual variant of T5 that was pre-trained on a new Common Crawl-based dataset covering $101$ languages.
We detail the design and modified training of mT5 and demonstrate its state-of-the-art performance on many multilingual benchmarks.
We also describe a simple technique to prevent ``accidental translation'' in the zero-shot setting, where a generative model chooses to (partially) translate its prediction into the wrong language.
All of the code and model checkpoints used in this work are publicly available.\footnote{\label{fn:code}\url{https://goo.gle/mt5-code}}
\end{abstract}

\section{Introduction}

Current natural language processing (NLP) pipelines often make use of transfer learning, where a model is pre-trained on a data-rich task before being fine-tuned on a downstream task of interest \citep{ruder2019transfer}.
The success of this paradigm is partially thanks to the release of parameter checkpoints for pre-trained models.
These checkpoints allow members of the NLP community to quickly attain strong performance on many tasks without needing to perform expensive pre-training themselves.
As one example, the pre-trained checkpoints for the ``Text-to-Text Transfer Transformer'' (T5) model released by \citet{2020t5} have been used to achieve state-of-the-art results on many benchmarks \citep[etc.]{khashabi2020unifiedqa,roberts2020much,kale2020text,izacard2020leveraging,nogueira2020document,narang2020wt5}.

Unfortunately, many of these language models were pre-trained solely on English-language text.
This significantly limits their use given that roughly 80\% of the world population does not speak English \citep{crystal2008two}.
One way the community has addressed this English-centricity has been to release dozens of models, each pre-trained on a single non-English language \citep[etc.]{carmo2020ptt5,de2019bertje,le2019flaubert,martin2019camembert,delobelle2020robbert,malmsten2020playing,nguyen2020phobert,polignano2019alberto}.
A more general solution is to produce multilingual models that have been pre-trained on a mixture of many languages.
Popular models of this type are mBERT \citep{devlin2018multilingual}, mBART \citep{liu2020multilingual}, and \mbox{XLM-R} \citep{conneau2019unsupervised}, which are multilingual variants of BERT \citep{devlin2018bert}, BART \citep{lewis2019bart}, and RoBERTa \citep{liu2019roberta}, respectively.

In this paper, we continue this tradition by releasing mT5, a multilingual variant of T5.
Our goal with mT5 is to produce a massively multilingual model that deviates as little as possible from the recipe used to create T5.
As such, mT5 inherits all of the benefits of T5 (described in \cref{sec:t5}), such as its general-purpose text-to-text format, its design based on insights from a large-scale empirical study, and its scale.
To train mT5, we introduce a multilingual variant of the C4 dataset called mC4.
mC4 comprises natural text in $101$ languages drawn from the public Common Crawl web scrape.
To validate the performance of mT5, we include results on several benchmark datasets, showing state-of-the-art results in many cases.
Finally, we characterize a problematic behavior of pre-trained generative multilingual language models in the zero-shot setting, where they erroneously translate part of their prediction into the wrong language.
To address this ``accidental translation'', we describe a simple procedure that involves mixing in unlabeled pre-training data during fine-tuning and demonstrate that it dramatically alleviates this issue.
We release our pre-trained models and code so that the community can leverage our work.\footnotemark[1]


\begin{figure*}
\centering
\includegraphics[width=\linewidth, trim=7 8 7 0, clip]{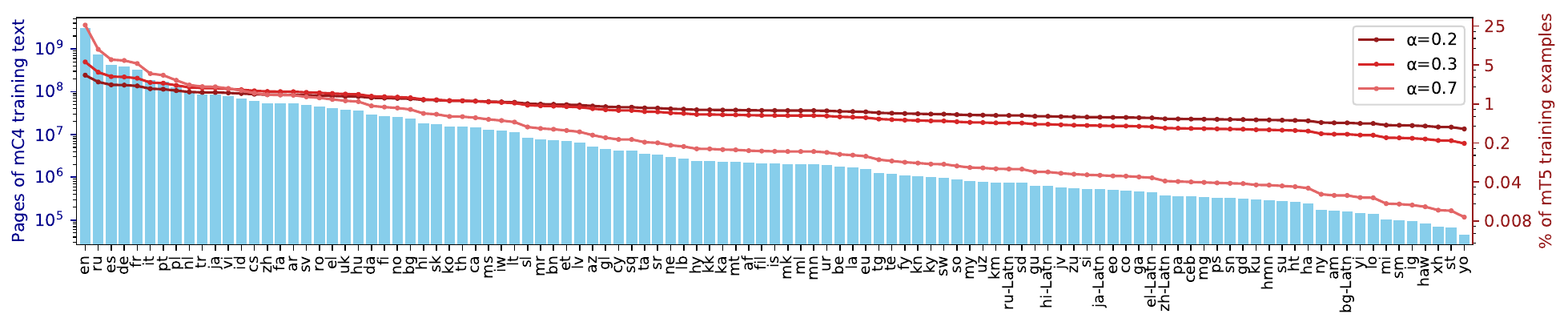}
\caption{Page counts per language in mC4 (left axis), and percentage of mT5 training examples coming from each language, for different language sampling exponents $\alpha$ (right axis). Our final model uses $\alpha$=0.3.}
\label{fig:corpus_stats}
\end{figure*}

\section{Background on T5 and C4}
\label{sec:t5}

In this section, we provide a short overview of T5 and the C4 pre-training dataset.
Further details are available in \citet{2020t5}.

T5 is a pre-trained language model whose primary distinction is its use of a unified ``text-to-text'' format for all text-based NLP problems.
This approach is natural for generative tasks (such as machine translation or abstractive summarization) where the task format requires the model to generate text conditioned on some input.
It is more unusual for classification tasks, where T5 is trained to output the literal text of the label (e.g.\ ``positive'' or ``negative'' for sentiment analysis) instead of a class index.
The primary advantage of this approach is that it allows the use of exactly the same training objective (teacher-forced maximum-likelihood) for every task, which in practice means that a single set of hyperparameters can be used for effective fine-tuning on any downstream task.
Similar unifying frameworks were proposed by \citet{keskar2019unifying} and \citet{mccann2018natural}.
Given the sequence-to-sequence structure of this task format, T5 uses a basic encoder-decoder Transformer architecture as originally proposed by \citet{vaswani2017attention}.
T5 is pre-trained on a masked language modeling ``span-corruption'' objective, where consecutive spans of input tokens are replaced with a mask token and the model is trained to reconstruct the masked-out tokens.

An additional distinguishing factor of T5 is its scale, with pre-trained model sizes available from 60 million to 11 billion parameters.
These models were pre-trained on around 1 trillion tokens of data.
Unlabeled data comes from the C4 dataset, which is a collection of about 750GB of English-language text sourced from the public Common Crawl web scrape.
C4 includes heuristics to extract only natural language (as opposed to boilerplate and other gibberish) in addition to extensive deduplication.
The pre-training objective, model architecture, scaling strategy, and many other design choices for T5 were chosen based on a large-scale empirical study described in detail in \citet{2020t5}.

\begin{table*}
\begin{center}
\footnotesize
\begin{tabular}{ lllll } 
 \toprule
 Model & Architecture & Parameters & \# languages & Data source \\
 \midrule
 mBERT \citep{devlin2018multilingual} & Encoder-only & $180$M & $104$ & Wikipedia \\
 XLM \citep{conneau2019cross} & Encoder-only & $570$M & $100$ & Wikipedia \\
 XLM-R \citep{conneau2019unsupervised} & Encoder-only & $270$M -- $550$M & $100$ & Common Crawl (CCNet) \\
 mBART \citep{lewis2019bart} & Encoder-decoder & $680$M & $25$ & Common Crawl (CC25) \\
 MARGE \citep{lewis2020pre} & Encoder-decoder & $960$M & $26$ & Wikipedia or CC-News \\
 mT5 (ours) & Encoder-decoder & $300$M -- $13$B & $101$ & Common Crawl (mC4) \\
 \bottomrule
\end{tabular}
\caption{Comparison of mT5 to existing massively multilingual pre-trained language models. Multiple versions of XLM and mBERT exist; we refer here to the ones that cover the most languages. Note that XLM-R counts five Romanized variants as separate languages, while we ignore six Romanized variants in the mT5 language count.}
\label{tab:comparison}
\end{center}
\end{table*}

\section{mC4 and mT5}

Our goal in this paper is to create a massively multilingual model that follows T5's recipe as closely as possible.
Towards this end, we develop an extended version of the C4 pre-training dataset that covers $101$ languages and introduce changes to T5 to better suit this multilinguality. 

\subsection{mC4}

The C4 dataset was explicitly designed to be English only: any page that was not given a probability of at least $99\%$ of being English by \texttt{langdetect}\footnote{\url{https://pypi.org/project/langdetect/}} was discarded.
In contrast, for mC4 we use \texttt{cld3}\footnote{\url{https://github.com/google/cld3}} to identify over $100$ languages.
Since some of these languages are relatively scarce on the internet, we make use of all of the $71$ monthly web scrapes released so far by Common Crawl.
This is dramatically more source data than was used for C4, for which the April 2019 web scrape alone was enough to provide plenty of English-language data.

An important heuristic filtering step in C4 was the removal of lines that did not end in an English terminal punctuation mark.
Since many languages do not use English terminal punctuation marks, we instead apply a ``line length filter'' that requires pages to contain at least three lines of text with 200 or more characters.
Otherwise, we follow C4's filtering by deduplicating lines across documents and removing pages containing bad words.\footnote{\url{https://github.com/LDNOOBW/}}
Finally, we detect each page's primary language using \texttt{cld3} and remove those with a confidence below 70\%.

After these filters are applied, we group the remaining pages by language and include in the corpus all languages with 10,000 or more pages. 
This produces text in 107 ``languages'' as defined by \texttt{cld3}.
However, we note that six of these are just script variants of the same spoken language (e.g.\ \texttt{ru} is Russian in Cyrillic script and \texttt{ru-Latn} is Russian in Latin script).
A histogram of the page counts for each language is shown in \cref{fig:corpus_stats}. Detailed dataset statistics including per-language token counts are shown in the appendix.

\subsection{mT5}

The model architecture and training procedure that we use for mT5 closely follows that of T5.
Specifically, we base mT5 on the ``T5.1.1'' recipe,\footnote{\url{https://github.com/google-research/text-to-text-transfer-transformer/blob/master/released\_checkpoints.md\#t511}} which improves upon T5 by using GeGLU nonlinearities \citep{shazeer2020glu}, scaling both $\mathrm{d_{model}}$ and $\mathrm{d_{ff}}$ instead of just $\mathrm{d_{ff}}$ in the larger models, and pre-training on unlabeled data only with no dropout.
We refer to \citet{2020t5} for further details on T5.

A major factor in pre-training multilingual models is how to sample data from each language.
Ultimately, this choice is a zero-sum game: If low-resource languages are sampled too often, the model may overfit; if high-resource languages are not trained on enough, the model will underfit.
We therefore take the approach used in \citep{devlin2018multilingual,conneau2019unsupervised,arivazhagan2019massively} and boost lower-resource languages by sampling examples according to the probability $p(L) \propto |L|^\alpha$, where $p(L)$ is the probability of sampling text from a given language during pre-training and $|L|$ is the number of examples in the language.
The hyperparameter $\alpha$ (typically with $\alpha < 1$) allows us to control how much to ``boost'' the probability of training on low-resource languages.
Values used by prior work include $\alpha = 0.7$ for mBERT \citep{devlin2018multilingual}, $\alpha = 0.3$ for XLM-R \citep{conneau2019unsupervised}, and $\alpha = 0.2$ for MMNMT \citep{arivazhagan2019massively}.
We tried all three of these values (ablation results in \cref{sec:ablation}) and found $\alpha = 0.3$ to give a reasonable compromise between performance on high- and low-resource languages.

The fact that our model covers over $100$ languages necessitates a larger vocabulary.
Following XLM-R \citep{conneau2018xnli}, we increase the vocabulary size to $250{,}000$ wordpieces.
As in T5, we use SentencePiece \citep{kudo2018sentencepiece,kudo2018subword} models trained with the language sampling rates used during pre-training.
To accommodate languages with large character sets like Chinese, we use a character coverage of $0.99999$ and enable SentencePiece's ``byte-fallback'' feature to ensure that any string can be uniquely encoded.

\begin{table*}[t!]
\centering
\footnotesize
\begin{tabular}{lccHcccp{7em}<{\centering\arraybackslash}}
\toprule
\multirow{2}{*}{Model} & \multicolumn{2}{c}{Sentence pair} & \multicolumn{2}{c}{Structured} & \multicolumn{3}{c}{Question answering} \\
\cmidrule(lr){2-3} \cmidrule(lr){4-5} \cmidrule(lr){6-8}
& XNLI & PAWS-X & POS & WikiAnn NER & XQuAD & MLQA & TyDiQA-GoldP \\
\midrule
Metrics & Acc. & Acc. & F1 & F1 & F1 / EM & F1 / EM & F1 / EM  \\
\midrule
\multicolumn{8}{l}{\emph{Cross-lingual zero-shot transfer (models fine-tuned on English data only)}} \\
\midrule
mBERT          & 65.4 & 81.9 & 70.3 & 62.2 & 64.5 / 49.4 & 61.4 / 44.2 & 59.7 / 43.9 \\
XLM            & 69.1 & 80.9 & 70.1 & 61.2 & 59.8 / 44.3 & 48.5 / 32.6 & 43.6 / 29.1 \\
InfoXLM                  & 81.4 & - & - & - & - / - & 73.6 / 55.2 & - / - \\
X-STILTs & 80.4 & 87.7 & 74.4 & 64.7 & 77.2 / 61.3 & 72.3 / 53.5 & 76.0 / 59.5 \\
XLM-R  & 79.2 & 86.4 & 72.6 & 65.4 & 76.6 / 60.8 & 71.6 / 53.2 & 65.1 / 45.0 \\
VECO & 79.9 & 88.7 & 75.1 & 65.7 & 77.3 / 61.8 & 71.7 / 53.2 & 67.6 / 49.1 \\
RemBERT & 80.8 & 87.5 & 76.5 & \textbf{70.1} & 79.6 / 64.0 & 73.1 / 55.0 & 77.0 / 63.0 \\
mT5-Small  & 67.5 & 82.4 & 72.6 & 50.5 & 58.1 / 42.5 & 54.6 / 37.1 & 35.2 / 23.2 \\
mT5-Base  & 75.4 & 86.4 & 72.6 & 55.7 & 67.0 / 49.0 & 64.6 / 45.0 & 57.2 / 41.2 \\
mT5-Large  & 81.1 & 88.9 & 72.6 & 58.5 & 77.8 / 61.5 & 71.2 / 51.7 & 69.9 / 52.2 \\
mT5-XL  & 82.9 & 89.6 & 72.6 & 65.5 & 79.5 / 63.6 & 73.5 / 54.5 & 75.9 / 59.4 \\
mT5-XXL  & \textbf{85.0} & \textbf{90.0} & 72.6 & 69.2 & \textbf{82.5 / 66.8} & \textbf{76.0 / 57.4} & \textbf{80.8 / 65.9} \\
\midrule
\multicolumn{8}{l}{\emph{Translate-train (models fine-tuned on English data plus translations in all target languages)}} \\
\midrule
XLM-R  & 82.6 & 90.4 &  -   & -    & 80.2 / 65.9 & 72.8 / 54.3 & 66.5 / 47.7 \\
\textsc{Filter} + Self-Teaching  & {83.9}   & 91.4 & 76.2 & - & {82.4} / {68.0}  & {76.2} / 57.7 & 68.3 / 50.9 \\
VECO & 83.0 & 91.1 & 75.1 & - & 79.9 / 66.3 & 73.1 / 54.9 & 75.0 / 58.9 \\
mT5-Small  & 64.7 & 79.9 & 72.6 & - & 64.3 / 49.5 & 56.6 / 38.8 & 48.2 / 34.0 \\
mT5-Base  & 75.9 & 89.3 & 72.6 & - & 75.3 / 59.7 & 67.6 / 48.5 & 64.0 / 47.7 \\
mT5-Large  & 81.8 & 91.2 & 72.6 & - & 81.2 / 65.9 & 73.9 / 55.2 & 71.1 / 54.9 \\
mT5-XL  & 84.8 & 91.0 & 72.6 & - & 82.7 / 68.1 & 75.1 / 56.6 & 79.9 / 65.3 \\
mT5-XXL  & \textbf{87.8} & \textbf{91.5} & 72.6 & - & \textbf{85.2 / 71.3} & \textbf{76.9 / 58.3} & \textbf{82.8 / 68.8} \\
\midrule
\multicolumn{8}{l}{\emph{In-language multitask (models fine-tuned on gold data in all target languages)}} \\
\midrule
mBERT & - & - & 91.5 & 89.1 & - & - & 77.6 / 68.0 \\
mT5-Small & - & - & - & 83.4 & - & - & 73.0 / 62.0 \\
mT5-Base & - & - & - & 85.4 & - & - & 80.8 / 70.0 \\
mT5-Large & - & - & - & 88.4 & - & - & 85.5 / 75.3 \\
mT5-XL & - & - & - & 90.9 & - & - & 87.5 / 78.1 \\
mT5-XXL & - & - & - & \textbf{91.2} & - & - & \textbf{88.5 / 79.1} \\
\bottomrule
\end{tabular}
\caption{Results on \textsc{xtreme} sentence-pair classification, structured prediction and question answering tasks. mBERT metrics are from \citet{hu2020xtreme}. Metrics for XLM, InfoXLM, X-STILTs and XLM-R are from \citet{fang2020filter}, though \citet{conneau2019unsupervised} report better performance of XLM-R on XNLI (80.9). All other metrics are from the original sources: \textsc{Filter} \cite{fang2020filter}, VECO \cite{luo2020veco} and RemBERT \cite{chung2020rethinking}. For the ``translate-train'' setting, we include English training data, so as to be comparable with \citet{fang2020filter} and \citet{luo2020veco}. This differs from the \textsc{xtreme} ``translate-train'' setup of \citet{hu2020xtreme}. For mT5 results on TyDi QA zero-shot, we report the median across five fine-tuning runs, as we observed high variance across runs. Full results for all languages in all tasks are provided in the appendix.
}

\label{tbl:detailed_results_full}
\end{table*}

\subsection{Comparison to related models}

To contextualize our new model, we provide a brief comparison with existing massively multilingual pre-trained language models.
For brevity, we focus on models that support more than a few dozen languages.
\Cref{tab:comparison} gives a high-level comparison of mT5 to the most similar models.

\textbf{mBERT} \citep{devlin2018multilingual} is a multilingual version of BERT \citep{devlin2018bert}.
Similar to our approach with mT5, mBERT follows the BERT recipe as closely as possible (same architecture, objective, etc.).
The primary difference is the training set: Instead of training on English Wikipedia and the Toronto Books Corpus, mBERT is trained on up to $104$ languages from Wikipedia.
\textbf{XLM} \citep{conneau2019cross} is also based on BERT but applies improved methods for pre-training multilingual language models including explicitly cross-lingual pre-training objectives.
Many pre-trained versions of XLM have been released; the most massively-multilingual variant was trained on $100$ languages from Wikipedia.
\textbf{XLM-R} \citep{conneau2019unsupervised} is an improved version of XLM based on the RoBERTa model \citep{liu2019roberta}.
\mbox{XLM-R} is trained with a cross-lingual masked language modeling objective on data in $100$ languages from Common Crawl.
To improve the pre-training data quality, pages from Common Crawl were filtered by an n-gram language model trained on Wikipedia \citep{wenzek2019ccnet}.
\textbf{mBART} \citep{liu2020multilingual} is a multilingual encoder-decoder model that is based on BART \citep{lewis2019bart}.
mBART is trained with a combination of span masking and sentence shuffling objectives on a subset of $25$ languages from the same data as XLM-R.
\textbf{MARGE} \citep{lewis2020pre} is a multilingual encoder-decoder model that is trained to reconstruct a document in one language by retrieving documents in other languages.
It uses data in $26$ languages from Wikipedia and CC-News \citep{liu2019roberta}.

\section{Experiments}

    To validate the performance of mT5, we evaluate our models on 6 tasks from the \textsc{xtreme} multilingual benchmark \citep{hu2020xtreme}: the XNLI \citep{conneau2018xnli} entailment task covering $14$ languages; the XQuAD \citep{Artetxe:etal:2019}, MLQA \citep{lewis2019mlqa}, and TyDi QA \citep{tydiqa} reading comprehension benchmarks with $10$, $7$, and $11$ languages respectively;
the Named Entity Recognition (NER) dataset of WikiAnn \citep{pan2017cross} restricted to the $40$ languages from \textsc{xtreme} \citep{hu2020xtreme}, and the \mbox{PAWS-X} \citep{yang2019paws} paraphrase identification dataset with $7$ languages.
We cast all tasks into the text-to-text format, i.e.\ generating the label text (XNLI and \mbox{PAWS-X}), entity tags and labels (WikiAnn NER), or answer (XQuAD, MLQA, and TyDi QA) directly in a generative fashion.
For NER, if there are multiple entities, then they are concatenated in the order they appear, and if there are no entities then the target text is ``None''.
We consider three variants of these tasks: (1)~``zero-shot'', where the model is fine-tuned only on English data, (2)~``translate-train'', adding machine translations from English into each target language, and (3)~``in-language multitask'', training on gold data in all target languages.
For brevity, we refer to \citet{hu2020xtreme} for further details on these benchmarks.

Following the original T5 recipe, we consider five model sizes: \textit{Small} ($\approx 300$M parameters), \textit{Base} ($580$M), \textit{Large} ($1.2$B), \textit{XL} ($3.7$B), and \textit{XXL} ($13$B).
The increase in parameter counts compared to the corresponding T5 model variants comes from the larger vocabulary used in mT5.
Note that, because mT5 is an encoder-decoder model, it has roughly twice as many parameters as correspondingly-sized encoder-only models such as XLM-R.
For example, the ``Large'' variant of XLM-R has $550$ million parameters whereas mT5-Large has around $1$ billion. 
However, the computational cost for text classification is roughly the same:
In both cases, the model processes a length-$T$ input sequence with an encoder of approximately equal size.
In an encoder-only model like XLM-R, the encoder processes one additional "CLS" token, which is used to generate the representation for classification.
In mT5, the decoder typically produces two additional tokens: the class label and an end-of-sequence token.
Since the decoder has the same architecture (ignoring encoder-decoder attention) as the encoder, the computational cost of classification with mT5 typically amounts to the cost of processing $T+2$ tokens compared to $T+1$ for an encoder-only model.
However, encoder-decoder architectures have the additional benefit of being applicable to generative tasks like abstractive summarization or dialog.

We pre-train our mT5 model variants for $1$ million steps on batches of $1024$ length-$1024$ input sequences, corresponding to roughly $1$ trillion input tokens total.
This is the same amount of pre-training as T5 and about $\frac{1}{6}$ as much as XLM-R.

We use the same inverse square-root learning rate schedule used by T5 during pre-training, with the learning rate set to $1 \big/ \sqrt{\max(n, k)}$ where $n$ is the current training iteration and $k = 10^{4}$ is the number of warm-up steps.
Following the T5.1.1 recipe, we do not apply dropout during pre-training.
We use the same self-supervised objective as T5, with $15$\% of tokens masked and an average noise span length of $3$.
We ablate some of these experimental details in \cref{sec:ablation}.

For fine-tuning, we use a constant learning rate of $0.001$ and dropout rate of $0.1$ for all tasks.
We use batch size $2^{17}$ for most tasks but increased this up to $2^{20}$ in a few cases based on performance on the validation set.
For early stopping, we save checkpoints every $200$ steps and choose the checkpoint with the highest validation performance.

\subsection{Results}

\Cref{tbl:detailed_results_full} presents our main results, with per-language breakdowns for each task given in the appendix.
Our largest model mT5-XXL exceeds state-of-the-art on all classification and QA tasks and is near SOTA on NER (69.2 vs.~70.1).
Note that unlike our model, InfoXLM \citep{chi2020infoxlm} and VECO \citep{luo2020veco} benefit from parallel training data, while X-STILTs \citep{phang2020english} leverages labeled data from tasks similar to the target task.
Overall, our results highlight the importance of model capacity in cross-lingual representation learning and suggest that scaling up a simple pre-training recipe can be a viable alternative to more complex techniques relying on LM filtering, parallel data, or intermediate tasks.

In the ``translate-train'' setting, we exceed state-of-the-art on all \textsc{xtreme} classification and QA tasks.
For these tasks, we fine-tune on the combination of the labeled English data and machine translations thereof.\footnote{We use the translation data provided by \citet{hu2020xtreme} throughout. On the PAWS-X task, \textsc{Filter} used translation data from the original task instead. Switching to this data would improve our scores slightly (mT5-XXL 91.5 $\rightarrow$ 92.0).}
This allows direct comparison with both \textsc{Filter} \citep{fang2020filter} as well as the XLM-R baseline of \citet{fang2020filter}.
Note that this setup differs from \textsc{xtreme} ``translate-train'' \citep{hu2020xtreme}, which excludes  English.

\begin{table}[t]
\centering
\footnotesize
\begin{tabular}{lcc}
\toprule
      & T5 & mT5 \\\midrule
Small & 87.2 / 79.1 & 84.7 / 76.4 \\
Base  & 92.1 / 85.4 & 89.6 / 83.8 \\
Large & 93.8 / 86.7 & 93.0 / 87.0 \\
XL    & 95.0 / 88.5 & 94.5 / 88.9 \\
XXL   & 96.2 / 91.3 & 95.6 / 90.4 \\ 
\bottomrule
\end{tabular}
\caption{Comparison of T5 vs.\ mT5 on SQuAD question answering (F1/EM).}
\label{tbl:squad}
\end{table}

\begin{figure}
    \centering
    \includegraphics[width=0.85\columnwidth]{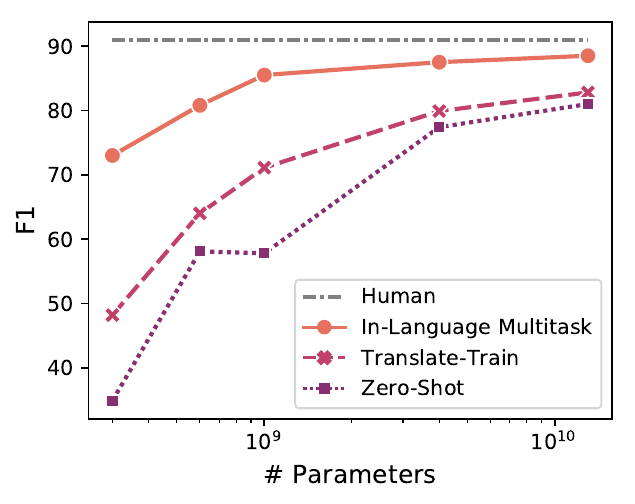}
    \caption{Average F1 on the TyDi QA GoldP task across languages. Performance improves with increasing model capacity. The importance of in-language training data (whether gold In-Lanugage Multitask or synthetic Translate-Train) decreases with model scale, as seen by Zero-Shot closing the quality gap.}
    \label{fig:tydiqa}
\end{figure}

Figure \ref{fig:tydiqa} shows that model capacity is key to improving performance on variants of the TyDi QA GoldP task in the absence of ``gold'' multilingual data:
For the smallest model, training on gold datasets (in-language multitask) achieves dramatically better performance than using weakly supervised data (translate-train) or English-only data (zero-shot), whereas the gap between these three settings is much smaller for the largest model.
For our two largest models, zero-shot and translate-train performance is nearly the same, showing that machine translations of the monolingual dataset bring diminishing returns as model capacity increases.
Overall, these trends point to the possibility of avoiding the costly step of annotating data in more than one language when using large models.

Massively multilingual models have been observed to underperform on a given language when compared to a similarly-sized ``dedicated'' model trained specifically for that language \citep{arivazhagan2019massively}.
To quantify this effect, we compare the performance of mT5 and T5 when fine-tuned on the SQuAD reading comprehension benchmark \citep{rajpurkar2016squad}.
The results are shown in \cref{tbl:squad}, with results for T5 reproduced from \citet{2020t5}.
While the \textit{Small} and \textit{Base} mT5 models fall short of their English T5 counterparts, we find that the larger models close the gap.
This suggests there may be a turning point past which the model has enough capacity to effectively learn $101$ languages without significant interference effects.

\subsection{Ablation} \label{sec:ablation}

We run six ablations, modifying various settings, using our \textit{Large} model as a baseline: (i) increase dropout to $0.1$ in hopes of mitigating overfitting on low-resource languages, (ii) decrease sequence length to $512$ (as was used in T5), (iii) increase the average noise span length in the pre-training objective to $10$ since we observe fewer characters per token than T5, (iv) adjust the language sampling exponent $\alpha$ to \{$0.2$, $0.7$\} as used in MMNMT \citep{arivazhagan2019massively} and mBERT \citep{devlin2018multilingual}, respectively, (v) turn off the ``line length filter'' in the mC4 data pipeline, and (vi) supplement mC4 with Wikipedia data\footnote{We use the 2020 Wikipedia data from TensorFlow Datasets, selecting the same languages as mBERT. \url{https://www.tensorflow.org/datasets/catalog/wikipedia}} from 103 languages.

\begin{table}[t]
\centering
\footnotesize
\begin{tabular}{lc}
\toprule
Model & Accuracy \\
\midrule
Baseline (mT5-Large) & \textbf{81.1} \\
Dropout $0.1$ & 77.6 \\
Sequence length $512$ & 80.5 \\
Span length $10$ & 78.6 \\
$\alpha = 0.7$ & 80.7 \\
$\alpha = 0.2$ & 80.7 \\
No line length filter & 79.1 \\
Add Wikipedia data & 80.3 \\
\bottomrule
\end{tabular}
\caption{Average XNLI zero-shot accuracy of various ablations on our mT5-Large model. Per-language metrics are shown in the appendix.}
\label{tbl:xnli_ablation_short}
\end{table}

The effect of these ablations on XNLI zero-shot accuracy is shown in \cref{tbl:xnli_ablation_short}.
In each case, the average XNLI score is lower than the mT5-Large baseline, justifying our chosen settings.
The line length filter provides a $+2$ point boost, corroborating the findings of \citet{conneau2019unsupervised} and \citet{2020t5} that filtering low-quality pages from Common Crawl is valuable.
Increasing the language sampling exponent $\alpha$ to $0.7$ has the expected effect of improving performance in high-resource languages (e.g.~Russian $81.5 \rightarrow 82.8$), while hurting low-resource languages (e.g.~Swahili $75.4 \rightarrow 70.6$), with the average effect being negative.
Conversely, lowering $\alpha$ to $0.2$ boosts one tail language slightly (Urdu $73.5 \rightarrow 73.9$) but is harmful elsewhere.
Detailed per-language metrics on XNLI and the results of our ablations on zero-shot XQuAD are provided in the appendix, showing similar trends.

\section{Zero-shot generation}

Since mT5 is a generative model, it can output arbitrary text predictions in a free form fashion.
This is in contrast to ``encoder-only'' models like mBERT and XLM(-R) that make a prediction by either extracting it from the input or producing a class label. 
We found that the lack of constraints during prediction caused mT5 to sometimes have trouble generating a well-formed prediction in a language unseen during fine-tuning.
Focusing on XQuAD zero-shot, we find that many of these errors are due to ``accidental translation'' into the fine-tuning language (English).
In this section, we characterize this behavior and demonstrate that it can be counteracted by mixing a small amount of our multilingual pre-training task into the fine-tuning stage.

\subsection{Illegal predictions}

In using a generative model for span selection (as in extractive QA tasks), we hope the model learns to generate \textbf{``legal'' spans} that are substrings of the provided context.
However, unlike encoder-based models like BERT, this is not a hard constraint of the model.
Notably, T5 learns to always output legal spans on SQuAD, suggesting this is not a major issue for generative models in simple cases.

A more challenging case for generative models is zero-shot cross-lingual span selection.
Here, a pre-trained multilingual model is fine-tuned on English but tested on other languages.
We want the model to generate legal non-English predictions despite having only seen English targets in fine-tuning.

In practice, while mT5 achieves SOTA on the zero-shot variants of XQuAD, MLQA and TyDi~QA, illegal predictions are still a problem.
For example, on zero-shot XQuAD, a non-trivial portion of mT5 mistakes are in fact illegal spans, for all model sizes (cf.~\cref{fig:dpt} ``Baseline'').
Through inspection, we find these illegal predictions mainly fall into three categories: (i)~normalization, (ii)~grammatical adjustment, and (iii)~accidental translation. \Cref{fig:illegal} provides examples of each type.

\begin{figure}
    \begin{subfigure}[b]{\columnwidth}
    \centering
    \includegraphics[width=\columnwidth, trim=7 9 10 7, clip]{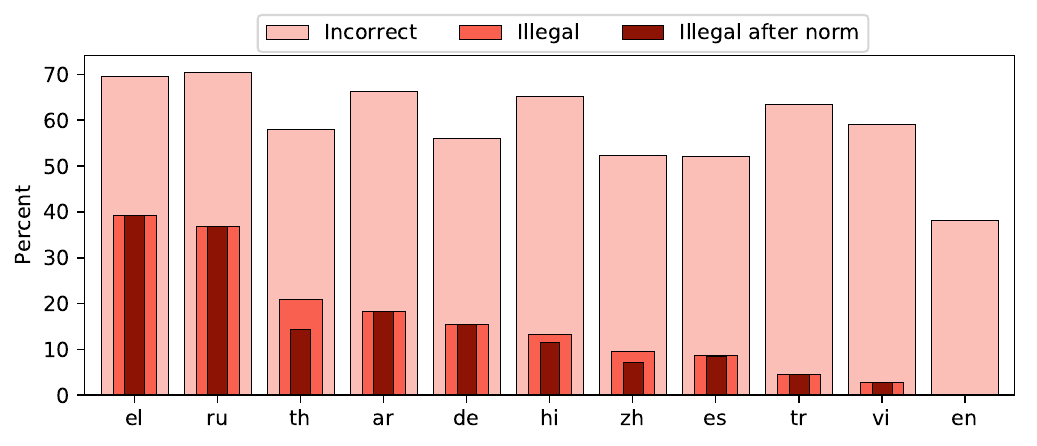}
    \caption{mT5-Small}
    \label{fig:xquad_errors_small}
    \end{subfigure}
    
    \vspace{4pt}
    
    \begin{subfigure}[b]{\columnwidth}
    \includegraphics[width=\columnwidth, trim=7 9 10 7, clip]{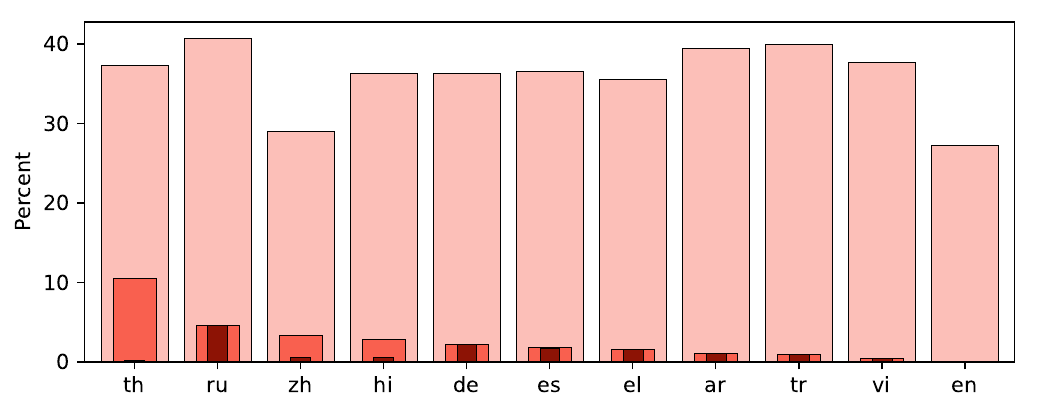}
    \caption{mT5-XXL}
    \label{fig:xquad_errors_xxl}
    \end{subfigure}
    \caption{Per-language error rates on XQuAD zero-shot, sorted by illegal rate. \textbf{Incorrect}: Not matching the target span. \textbf{Illegal}: Missing from the input context. \textbf{Illegal after norm}: Illegal even after Unicode NFKC normalization is applied to the prediction and context.}
    \label{fig:xquad_errors}
\end{figure}

\textbf{Normalization} indicates predictions that would be legal, except that ``equivalent'' Unicode characters have been substituted, so a legal span may be recovered through Unicode NFKC normalization.
This is particularly common in Thai, Chinese and Hindi, where most mT5-XXL illegal predictions are resolved by normalization, as seen in \cref{fig:xquad_errors_xxl}.

\begin{table}
    \centering
    \includegraphics[width=\columnwidth, trim=6 1 3 0, clip]{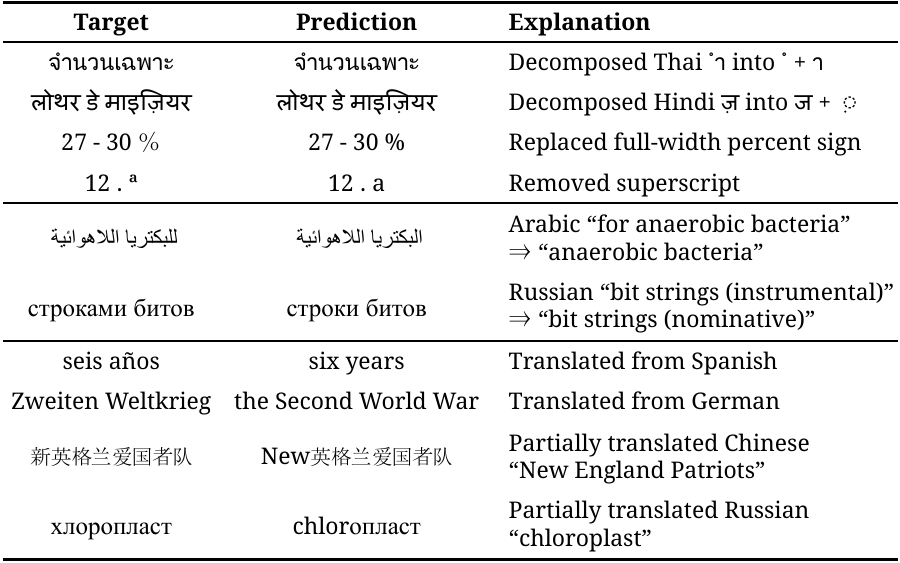}
    \caption{Illegal mT5-XXL predictions on XQuAD zero-shot, illustrating normalization (top), grammatical adjustment (middle) and translation (bottom).}
    \label{fig:illegal}
\end{table}

\textbf{Grammatical adjustment} involves minor morphological changes to the original text. 
We frequently observe these adjustments when the target span cannot stand as a well-formed answer on its own.
For example, mT5-XXL's Arabic and Russian predictions in the middle rows of \cref{fig:illegal} are judged by native speakers as correct and grammatical answers to the posed XQuAD questions, while the gold targets are judged as ungrammatical answers.
This type of illegal prediction is most common in languages with extensive grammatical case marking, such as Russian, Turkish and German.


\textbf{Accidental translation} involves the model translating part or all of a contextual span into English (the language of all fine-tuning data).
On the one hand, it is remarkable that mT5 performs ``spontaneous'' translation despite never seeing parallel training data.
On the other, as practitioners we would ideally be able to control this behavior.


We observe accidental translation across all model sizes and all XQuAD languages.
The problem is most prevalent in mT5-Small and mT5-Base, where from manual inspection, half or more of the illegal predictions within each language exhibit accidental translation, with many of the illegal predictions coming from Greek and Russian, as shown in \cref{fig:xquad_errors_small}.
While we do observe full phrase translations, a more common occurrence is \emph{partial} translation, where the model outputs a token or two of English before reverting to the correct target language.
The transition may even occur mid-word, as in the prediction ``chlor\foreignlanguage{russian}{опласт}'', where the first half of the target ``\foreignlanguage{russian}{хлоропласт}'' (Russian: chloroplast) has been translated to English.


\subsection{Preventing accidental translation}








The most direct solution to avoiding accidental translation on span selection tasks would be to modify our inference procedure.
As is common practice with encoder-based models, we could devise a task-specific fine-tuning mechanism that restricts the model to perform ranking over legal spans, removing the possibility of illegal predictions entirely.
While this would likely improve our zero-shot metrics, it is unsatisfying for two reasons:
First, it implies taking a step backward from the general text-to-text interface, as different tasks would demand different types of inference.
Second, this solution won't extend to more ``open-ended'' zero-shot generative tasks like summarization, where the legal output space can't be easily delimited. 


For these reasons, we consider a more general solution that remains within the text-to-text framework and can apply to all zero-shot generation tasks.
Our motivating intuition is that the reason the model outputs English when given a non-English test input is that it has never observed a non-English target during fine-tuning.
As English-only fine-tuning proceeds, the model's assigned likelihood of non-English tokens presumably decreases, eventually reaching the point where English becomes the most likely answer to any question.

To prevent the model from ``forgetting'' how to generate other languages, we use a strategy inspired by domain/task-adaptive pre-training \cite{howard2018universal,gururangan2020don}:
We simply mix in our unsupervised multilingual pre-training task during fine-tuning.
A similar approach was explored by \citet{liu2020exploring}.
We use the same mC4 task definition as in pre-training, with two adjustments:
First, we remove all ``sentinel'' tokens (corresponding to non-masked spans in the input text) from the target sequence, as otherwise we observe occasional sentinels in downstream predictions.
Second, we reduce the language sampling parameter $\alpha$ from $0.3$ to $0.1$.
This produces a near-uniform distribution of languages, encouraging the model to treat all languages as equally likely.\footnote{Alternatively, one could mix in unlabeled data only for a single language at a time. However, we believe this is contrary to the spirit of multilingual models and zero-shot evaluation.}

\begin{figure}
    \centering
    \includegraphics[width=\columnwidth, trim=7 7 7 7, clip]{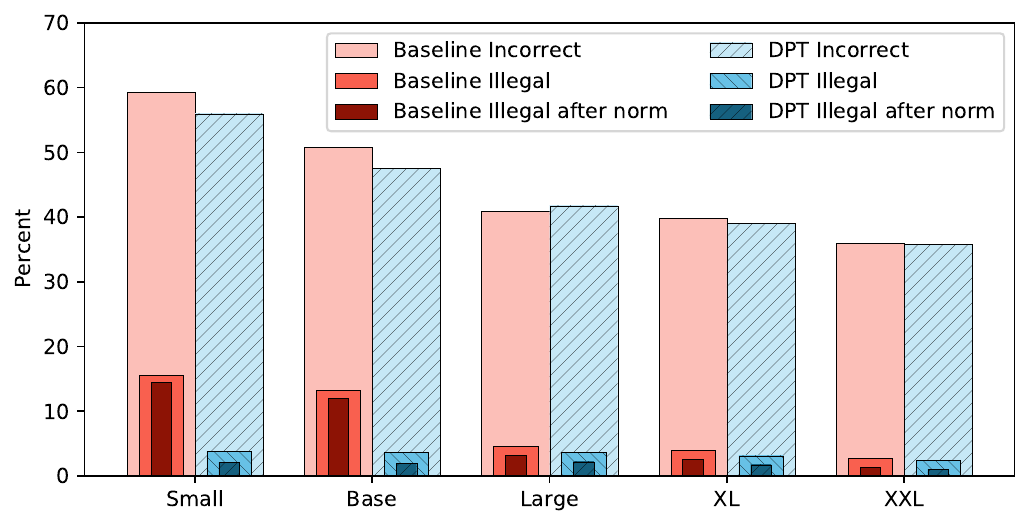}
    \caption{Error rates of mT5 on XQuAD zero-shot. \textbf{Baseline}: Fine-tuning on XQuAD alone. \textbf{Domain Preserving Training (DPT)}: Mixing in the unsupervised mC4 task with fine-tuning.}
    \label{fig:dpt}
\end{figure}

With these changes, we mix a small amount of our unsupervised task (covering 101 languages) into XQuAD fine-tuning, at a ratio of just $1$:$100$. \Cref{fig:dpt} shows the results on XQuAD zero-shot error rates.
The addition of even this small amount of multilingual data has a marked effect on the mT5-Small and mT5-Base models (where accidental translation was most rampant), reducing the illegal prediction rates by more than $70\%$ (relative), and contributing to an overall reduction in errors.

\section{Conclusion}

In this paper, we introduced mT5 and mC4: massively multilingual variants of the T5 model and C4 dataset.
We demonstrated that the T5 recipe is straightforwardly applicable to the multilingual setting, and achieved strong performance on a diverse set of benchmarks.
We also characterized illegal predictions that can occur in zero-shot evaluation of multilingual pre-trained generative models, and described a simple technique to avoid this issue.
We release all code and pre-trained datasets used in this paper to facilitate future work on multilingual language understanding.\footnote{\url{https://goo.gle/mt5-code}}

\subsection*{Acknowledgements}

We thank Melvin Johnson for tips on the translate-train procedure for \textsc{xtreme} and Itai Rolnick for help with infrastructure.

\bibliography{naaclhlt2021}

\begin{thebibliography}{49}
\expandafter\ifx\csname natexlab\endcsname\relax\def\natexlab#1{#1}\fi

\bibitem[{Arivazhagan et~al.(2019)Arivazhagan, Bapna, Firat, Lepikhin, Johnson,
  Krikun, Chen, Cao, Foster, Cherry et~al.}]{arivazhagan2019massively}
Naveen Arivazhagan, Ankur Bapna, Orhan Firat, Dmitry Lepikhin, Melvin Johnson,
  Maxim Krikun, Mia~Xu Chen, Yuan Cao, George Foster, Colin Cherry, et~al.
  2019.
\newblock Massively multilingual neural machine translation in the wild:
  Findings and challenges.
\newblock \emph{arXiv preprint arXiv:1907.05019}.

\bibitem[{Artetxe et~al.(2020)Artetxe, Ruder, and Yogatama}]{Artetxe:etal:2019}
Mikel Artetxe, Sebastian Ruder, and Dani Yogatama. 2020.
\newblock \href {https://doi.org/10.18653/v1/2020.acl-main.421} {On the
  cross-lingual transferability of monolingual representations}.
\newblock In \emph{Proceedings of the 58th Annual Meeting of the Association
  for Computational Linguistics}, pages 4623--4637, Online. Association for
  Computational Linguistics.

\bibitem[{Carmo et~al.(2020)Carmo, Piau, Campiotti, Nogueira, and
  Lotufo}]{carmo2020ptt5}
Diedre Carmo, Marcos Piau, Israel Campiotti, Rodrigo Nogueira, and Roberto
  Lotufo. 2020.
\newblock {PTT5}: Pretraining and validating the t5 model on brazilian
  portuguese data.
\newblock \emph{arXiv preprint arXiv:2008.09144}.

\bibitem[{Chi et~al.(2020)Chi, Dong, Wei, Yang, Singhal, Wang, Song, Mao,
  Huang, and Zhou}]{chi2020infoxlm}
Zewen Chi, Li~Dong, Furu Wei, Nan Yang, Saksham Singhal, Wenhui Wang, Xia Song,
  Xian-Ling Mao, Heyan Huang, and Ming Zhou. 2020.
\newblock {InfoXLM}: An information-theoretic framework for cross-lingual
  language model pre-training.
\newblock \emph{arXiv preprint arXiv:2007.07834}.

\bibitem[{Chung et~al.(2020)Chung, F{\'e}vry, Tsai, Johnson, and
  Ruder}]{chung2020rethinking}
Hyung~Won Chung, Thibault F{\'e}vry, Henry Tsai, Melvin Johnson, and Sebastian
  Ruder. 2020.
\newblock Rethinking embedding coupling in pre-trained language models.
\newblock \emph{arXiv preprint arXiv:2010.12821}.

\bibitem[{Clark et~al.(2020)Clark, Choi, Collins, Garrette, Kwiatkowski,
  Nikolaev, and Palomaki}]{tydiqa}
Jonathan~H. Clark, Eunsol Choi, Michael Collins, Dan Garrette, Tom Kwiatkowski,
  Vitaly Nikolaev, and Jennimaria Palomaki. 2020.
\newblock \href {https://doi.org/10.1162/tacl\_a\_00317} {{TyDi QA}: A
  benchmark for information-seeking question answering in typologically diverse
  languages}.
\newblock \emph{Transactions of the Association for Computational Linguistics},
  8:454--470.

\bibitem[{Conneau et~al.(2020)Conneau, Khandelwal, Goyal, Chaudhary, Wenzek,
  Guzm{\'a}n, Grave, Ott, Zettlemoyer, and Stoyanov}]{conneau2019unsupervised}
Alexis Conneau, Kartikay Khandelwal, Naman Goyal, Vishrav Chaudhary, Guillaume
  Wenzek, Francisco Guzm{\'a}n, Edouard Grave, Myle Ott, Luke Zettlemoyer, and
  Veselin Stoyanov. 2020.
\newblock \href {https://doi.org/10.18653/v1/2020.acl-main.747} {Unsupervised
  cross-lingual representation learning at scale}.
\newblock In \emph{Proceedings of the 58th Annual Meeting of the Association
  for Computational Linguistics}, pages 8440--8451, Online. Association for
  Computational Linguistics.

\bibitem[{Conneau and Lample(2019)}]{conneau2019cross}
Alexis Conneau and Guillaume Lample. 2019.
\newblock \href
  {https://proceedings.neurips.cc/paper/2019/file/c04c19c2c2474dbf5f7ac4372c5b9af1-Paper.pdf}
  {Cross-lingual language model pretraining}.
\newblock In \emph{Advances in Neural Information Processing Systems},
  volume~32, pages 7059--7069.

\bibitem[{Conneau et~al.(2018)Conneau, Rinott, Lample, Williams, Bowman,
  Schwenk, and Stoyanov}]{conneau2018xnli}
Alexis Conneau, Ruty Rinott, Guillaume Lample, Adina Williams, Samuel Bowman,
  Holger Schwenk, and Veselin Stoyanov. 2018.
\newblock \href {https://doi.org/10.18653/v1/D18-1269} {{XNLI}: Evaluating
  cross-lingual sentence representations}.
\newblock In \emph{Proceedings of the 2018 Conference on Empirical Methods in
  Natural Language Processing}, pages 2475--2485, Brussels, Belgium.
  Association for Computational Linguistics.

\bibitem[{Crystal(2008)}]{crystal2008two}
David Crystal. 2008.
\newblock Two thousand million?
\newblock \emph{English today}, 24(1):3--6.

\bibitem[{de~Vries et~al.(2019)de~Vries, van Cranenburgh, Bisazza, Caselli, van
  Noord, and Nissim}]{de2019bertje}
Wietse de~Vries, Andreas van Cranenburgh, Arianna Bisazza, Tommaso Caselli,
  Gertjan van Noord, and Malvina Nissim. 2019.
\newblock {BERTje}: A dutch {BERT} model.
\newblock \emph{arXiv preprint arXiv:1912.09582}.

\bibitem[{Delobelle et~al.(2020)Delobelle, Winters, and
  Berendt}]{delobelle2020robbert}
Pieter Delobelle, Thomas Winters, and Bettina Berendt. 2020.
\newblock {RobBERT}: a dutch {RoBERTa}-based language model.
\newblock \emph{arXiv preprint arXiv:2001.06286}.

\bibitem[{Devlin(2018)}]{devlin2018multilingual}
Jacob Devlin. 2018.
\newblock {Multilingual BERT README}.
\newblock
  \url{https://github.com/google-research/bert/blob/master/multilingual.md}.

\bibitem[{Devlin et~al.(2019)Devlin, Chang, Lee, and
  Toutanova}]{devlin2018bert}
Jacob Devlin, Ming-Wei Chang, Kenton Lee, and Kristina Toutanova. 2019.
\newblock \href {https://doi.org/10.18653/v1/N19-1423} {{BERT}: Pre-training of
  deep bidirectional transformers for language understanding}.
\newblock In \emph{Proceedings of the 2019 Conference of the North {A}merican
  Chapter of the Association for Computational Linguistics: Human Language
  Technologies, Volume 1 (Long and Short Papers)}, pages 4171--4186,
  Minneapolis, Minnesota. Association for Computational Linguistics.

\bibitem[{Fang et~al.(2020)Fang, Wang, Gan, Sun, and Liu}]{fang2020filter}
Yuwei Fang, Shuohang Wang, Zhe Gan, Siqi Sun, and Jingjing Liu. 2020.
\newblock {FILTER}: An enhanced fusion method for cross-lingual language
  understanding.
\newblock \emph{arXiv preprint arXiv:2009.05166}.

\bibitem[{Gururangan et~al.(2020)Gururangan, Marasovi{\'c}, Swayamdipta, Lo,
  Beltagy, Downey, and Smith}]{gururangan2020don}
Suchin Gururangan, Ana Marasovi{\'c}, Swabha Swayamdipta, Kyle Lo, Iz~Beltagy,
  Doug Downey, and Noah~A. Smith. 2020.
\newblock \href {https://doi.org/10.18653/v1/2020.acl-main.740} {Don{'}t stop
  pretraining: Adapt language models to domains and tasks}.
\newblock In \emph{Proceedings of the 58th Annual Meeting of the Association
  for Computational Linguistics}, pages 8342--8360, Online. Association for
  Computational Linguistics.

\bibitem[{Howard and Ruder(2018)}]{howard2018universal}
Jeremy Howard and Sebastian Ruder. 2018.
\newblock \href {https://doi.org/10.18653/v1/P18-1031} {Universal language
  model fine-tuning for text classification}.
\newblock In \emph{Proceedings of the 56th Annual Meeting of the Association
  for Computational Linguistics (Volume 1: Long Papers)}, pages 328--339,
  Melbourne, Australia. Association for Computational Linguistics.

\bibitem[{Hu et~al.(2020)Hu, Ruder, Siddhant, Neubig, Firat, and
  Johnson}]{hu2020xtreme}
Junjie Hu, Sebastian Ruder, Aditya Siddhant, Graham Neubig, Orhan Firat, and
  Melvin Johnson. 2020.
\newblock {XTREME}: A massively multilingual multi-task benchmark for
  evaluating cross-lingual generalization.
\newblock \emph{arXiv preprint arXiv:2003.11080}.

\bibitem[{Izacard and Grave(2020)}]{izacard2020leveraging}
Gautier Izacard and Edouard Grave. 2020.
\newblock Leveraging passage retrieval with generative models for open domain
  question answering.
\newblock \emph{arXiv preprint arXiv:2007.01282}.

\bibitem[{Kale(2020)}]{kale2020text}
Mihir Kale. 2020.
\newblock Text-to-text pre-training for data-to-text tasks.
\newblock \emph{arXiv preprint arXiv:2005.10433}.

\bibitem[{Keskar et~al.(2019)Keskar, McCann, Xiong, and
  Socher}]{keskar2019unifying}
Nitish~Shirish Keskar, Bryan McCann, Caiming Xiong, and Richard Socher. 2019.
\newblock Unifying question answering and text classification via span
  extraction.
\newblock \emph{arXiv preprint arXiv:1904.09286}.

\bibitem[{Khashabi et~al.(2020)Khashabi, Min, Khot, Sabharwal, Tafjord, Clark,
  and Hajishirzi}]{khashabi2020unifiedqa}
Daniel Khashabi, Sewon Min, Tushar Khot, Ashish Sabharwal, Oyvind Tafjord,
  Peter Clark, and Hannaneh Hajishirzi. 2020.
\newblock \href {https://www.aclweb.org/anthology/2020.findings-emnlp.171}
  {{UnifiedQA}: Crossing format boundaries with a single {QA} system}.
\newblock In \emph{Findings of the Association for Computational Linguistics:
  EMNLP 2020}, pages 1896--1907, Online. Association for Computational
  Linguistics.

\bibitem[{Kudo(2018)}]{kudo2018subword}
Taku Kudo. 2018.
\newblock \href {https://doi.org/10.18653/v1/P18-1007} {Subword regularization:
  Improving neural network translation models with multiple subword
  candidates}.
\newblock In \emph{Proceedings of the 56th Annual Meeting of the Association
  for Computational Linguistics (Volume 1: Long Papers)}, pages 66--75,
  Melbourne, Australia. Association for Computational Linguistics.

\bibitem[{Kudo and Richardson(2018)}]{kudo2018sentencepiece}
Taku Kudo and John Richardson. 2018.
\newblock \href {https://doi.org/10.18653/v1/D18-2012} {{S}entence{P}iece: A
  simple and language independent subword tokenizer and detokenizer for neural
  text processing}.
\newblock In \emph{Proceedings of the 2018 Conference on Empirical Methods in
  Natural Language Processing: System Demonstrations}, pages 66--71, Brussels,
  Belgium. Association for Computational Linguistics.

\bibitem[{Le et~al.(2020)Le, Vial, Frej, Segonne, Coavoux, Lecouteux, Allauzen,
  Crabb{\'e}, Besacier, and Schwab}]{le2019flaubert}
Hang Le, Lo{\"\i}c Vial, Jibril Frej, Vincent Segonne, Maximin Coavoux,
  Benjamin Lecouteux, Alexandre Allauzen, Benoit Crabb{\'e}, Laurent Besacier,
  and Didier Schwab. 2020.
\newblock \href {https://www.aclweb.org/anthology/2020.lrec-1.302}
  {{F}lau{BERT}: Unsupervised language model pre-training for {F}rench}.
\newblock In \emph{Proceedings of the 12th Language Resources and Evaluation
  Conference}, pages 2479--2490, Marseille, France. European Language Resources
  Association.

\bibitem[{Lewis et~al.(2020{\natexlab{a}})Lewis, Ghazvininejad, Ghosh,
  Aghajanyan, Wang, and Zettlemoyer}]{lewis2020pre}
Mike Lewis, Marjan Ghazvininejad, Gargi Ghosh, Armen Aghajanyan, Sida Wang, and
  Luke Zettlemoyer. 2020{\natexlab{a}}.
\newblock Pre-training via paraphrasing.
\newblock \emph{arXiv preprint arXiv:2006.15020}.

\bibitem[{Lewis et~al.(2020{\natexlab{b}})Lewis, Liu, Goyal, Ghazvininejad,
  Mohamed, Levy, Stoyanov, and Zettlemoyer}]{lewis2019bart}
Mike Lewis, Yinhan Liu, Naman Goyal, Marjan Ghazvininejad, Abdelrahman Mohamed,
  Omer Levy, Veselin Stoyanov, and Luke Zettlemoyer. 2020{\natexlab{b}}.
\newblock \href {https://doi.org/10.18653/v1/2020.acl-main.703} {{BART}:
  Denoising sequence-to-sequence pre-training for natural language generation,
  translation, and comprehension}.
\newblock In \emph{Proceedings of the 58th Annual Meeting of the Association
  for Computational Linguistics}, pages 7871--7880, Online. Association for
  Computational Linguistics.

\bibitem[{Lewis et~al.(2019)Lewis, O{\u{g}}uz, Rinott, Riedel, and
  Schwenk}]{lewis2019mlqa}
Patrick Lewis, Barlas O{\u{g}}uz, Ruty Rinott, Sebastian Riedel, and Holger
  Schwenk. 2019.
\newblock {MLQA}: Evaluating cross-lingual extractive question answering.
\newblock \emph{arXiv preprint arXiv:1910.07475}.

\bibitem[{Liu et~al.(2020{\natexlab{a}})Liu, Gu, Goyal, Li, Edunov,
  Ghazvininejad, Lewis, and Zettlemoyer}]{liu2020multilingual}
Yinhan Liu, Jiatao Gu, Naman Goyal, Xian Li, Sergey Edunov, Marjan
  Ghazvininejad, Mike Lewis, and Luke Zettlemoyer. 2020{\natexlab{a}}.
\newblock Multilingual denoising pre-training for neural machine translation.
\newblock \emph{arXiv preprint arXiv:2001.08210}.

\bibitem[{Liu et~al.(2019)Liu, Ott, Goyal, Du, Joshi, Chen, Levy, Lewis,
  Zettlemoyer, and Stoyanov}]{liu2019roberta}
Yinhan Liu, Myle Ott, Naman Goyal, Jingfei Du, Mandar Joshi, Danqi Chen, Omer
  Levy, Mike Lewis, Luke Zettlemoyer, and Veselin Stoyanov. 2019.
\newblock {RoBERTa}: A robustly optimized {BERT} pretraining approach.
\newblock \emph{arXiv preprint arXiv:1907.11692}.

\bibitem[{Liu et~al.(2020{\natexlab{b}})Liu, Winata, Madotto, and
  Fung}]{liu2020exploring}
Zihan Liu, Genta~Indra Winata, Andrea Madotto, and Pascale Fung.
  2020{\natexlab{b}}.
\newblock Exploring fine-tuning techniques for pre-trained cross-lingual models
  via continual learning.
\newblock \emph{arXiv preprint arXiv:2004.14218}.

\bibitem[{Luo et~al.(2020)Luo, Wang, Liu, Liu, Bi, Huang, Huang, and
  Si}]{luo2020veco}
Fuli Luo, Wei Wang, Jiahao Liu, Yijia Liu, Bin Bi, Songfang Huang, Fei Huang,
  and Luo Si. 2020.
\newblock Veco: Variable encoder-decoder pre-training for cross-lingual
  understanding and generation.
\newblock \emph{arXiv preprint arXiv:2010.16046}.

\bibitem[{Malmsten et~al.(2020)Malmsten, B{\"o}rjeson, and
  Haffenden}]{malmsten2020playing}
Martin Malmsten, Love B{\"o}rjeson, and Chris Haffenden. 2020.
\newblock Playing with words at the national library of sweden--making a
  swedish {BERT}.
\newblock \emph{arXiv preprint arXiv:2007.01658}.

\bibitem[{Martin et~al.(2020)Martin, Muller, Ortiz~Su{\'a}rez, Dupont, Romary,
  de~la Clergerie, Seddah, and Sagot}]{martin2019camembert}
Louis Martin, Benjamin Muller, Pedro~Javier Ortiz~Su{\'a}rez, Yoann Dupont,
  Laurent Romary, {\'E}ric de~la Clergerie, Djam{\'e} Seddah, and Beno{\^\i}t
  Sagot. 2020.
\newblock \href {https://doi.org/10.18653/v1/2020.acl-main.645} {{C}amem{BERT}:
  a tasty {F}rench language model}.
\newblock In \emph{Proceedings of the 58th Annual Meeting of the Association
  for Computational Linguistics}, pages 7203--7219, Online. Association for
  Computational Linguistics.

\bibitem[{McCann et~al.(2018)McCann, Keskar, Xiong, and
  Socher}]{mccann2018natural}
Bryan McCann, Nitish~Shirish Keskar, Caiming Xiong, and Richard Socher. 2018.
\newblock The natural language decathlon: Multitask learning as question
  answering.
\newblock \emph{arXiv preprint arXiv:1806.08730}.

\bibitem[{Narang et~al.(2020)Narang, Raffel, Lee, Roberts, Fiedel, and
  Malkan}]{narang2020wt5}
Sharan Narang, Colin Raffel, Katherine Lee, Adam Roberts, Noah Fiedel, and
  Karishma Malkan. 2020.
\newblock {WT5}?! {T}raining text-to-text models to explain their predictions.
\newblock \emph{arXiv preprint arXiv:2004.14546}.

\bibitem[{Nguyen and Tuan~Nguyen(2020)}]{nguyen2020phobert}
Dat~Quoc Nguyen and Anh Tuan~Nguyen. 2020.
\newblock \href {https://www.aclweb.org/anthology/2020.findings-emnlp.92}
  {{P}ho{BERT}: Pre-trained language models for {V}ietnamese}.
\newblock In \emph{Findings of the Association for Computational Linguistics:
  EMNLP 2020}, pages 1037--1042, Online. Association for Computational
  Linguistics.

\bibitem[{Nogueira et~al.(2020)Nogueira, Jiang, Pradeep, and
  Lin}]{nogueira2020document}
Rodrigo Nogueira, Zhiying Jiang, Ronak Pradeep, and Jimmy Lin. 2020.
\newblock \href {https://www.aclweb.org/anthology/2020.findings-emnlp.63}
  {Document ranking with a pretrained sequence-to-sequence model}.
\newblock In \emph{Findings of the Association for Computational Linguistics:
  EMNLP 2020}, pages 708--718, Online. Association for Computational
  Linguistics.

\bibitem[{Pan et~al.(2017)Pan, Zhang, May, Nothman, Knight, and
  Ji}]{pan2017cross}
Xiaoman Pan, Boliang Zhang, Jonathan May, Joel Nothman, Kevin Knight, and Heng
  Ji. 2017.
\newblock \href {https://doi.org/10.18653/v1/P17-1178} {Cross-lingual name
  tagging and linking for 282 languages}.
\newblock In \emph{Proceedings of the 55th Annual Meeting of the Association
  for Computational Linguistics (Volume 1: Long Papers)}, pages 1946--1958,
  Vancouver, Canada. Association for Computational Linguistics.

\bibitem[{Phang et~al.(2020)Phang, Htut, Pruksachatkun, Liu, Vania, Kann,
  Calixto, and Bowman}]{phang2020english}
Jason Phang, Phu~Mon Htut, Yada Pruksachatkun, Haokun Liu, Clara Vania,
  Katharina Kann, Iacer Calixto, and Samuel~R Bowman. 2020.
\newblock English intermediate-task training improves zero-shot cross-lingual
  transfer too.
\newblock \emph{arXiv preprint arXiv:2005.13013}.

\bibitem[{Polignano et~al.(2019)Polignano, Basile, de~Gemmis, Semeraro, and
  Basile}]{polignano2019alberto}
Marco Polignano, Pierpaolo Basile, Marco de~Gemmis, Giovanni Semeraro, and
  Valerio Basile. 2019.
\newblock {AlBERTo}: Italian {BERT} language understanding model for {NLP}
  challenging tasks based on tweets.
\newblock In \emph{CLiC-it}.

\bibitem[{Raffel et~al.(2020)Raffel, Shazeer, Roberts, Lee, Narang, Matena,
  Zhou, Li, and Liu}]{2020t5}
Colin Raffel, Noam Shazeer, Adam Roberts, Katherine Lee, Sharan Narang, Michael
  Matena, Yanqi Zhou, Wei Li, and Peter~J. Liu. 2020.
\newblock \href {http://jmlr.org/papers/v21/20-074.html} {Exploring the limits
  of transfer learning with a unified text-to-text transformer}.
\newblock \emph{Journal of Machine Learning Research}, 21(140):1--67.

\bibitem[{Rajpurkar et~al.(2016)Rajpurkar, Zhang, Lopyrev, and
  Liang}]{rajpurkar2016squad}
Pranav Rajpurkar, Jian Zhang, Konstantin Lopyrev, and Percy Liang. 2016.
\newblock \href {https://doi.org/10.18653/v1/D16-1264} {{SQ}u{AD}: 100,000+
  questions for machine comprehension of text}.
\newblock In \emph{Proceedings of the 2016 Conference on Empirical Methods in
  Natural Language Processing}, pages 2383--2392, Austin, Texas. Association
  for Computational Linguistics.

\bibitem[{Roberts et~al.(2020)Roberts, Raffel, and Shazeer}]{roberts2020much}
Adam Roberts, Colin Raffel, and Noam Shazeer. 2020.
\newblock \href {https://www.aclweb.org/anthology/2020.emnlp-main.437} {How
  much knowledge can you pack into the parameters of a language model?}
\newblock In \emph{Proceedings of the 2020 Conference on Empirical Methods in
  Natural Language Processing (EMNLP)}, pages 5418--5426, Online. Association
  for Computational Linguistics.

\bibitem[{Ruder et~al.(2019)Ruder, Peters, Swayamdipta, and
  Wolf}]{ruder2019transfer}
Sebastian Ruder, Matthew~E. Peters, Swabha Swayamdipta, and Thomas Wolf. 2019.
\newblock \href {https://doi.org/10.18653/v1/N19-5004} {Transfer learning in
  natural language processing}.
\newblock In \emph{Proceedings of the 2019 Conference of the North {A}merican
  Chapter of the Association for Computational Linguistics: Tutorials}, pages
  15--18, Minneapolis, Minnesota. Association for Computational Linguistics.

\bibitem[{Shazeer(2020)}]{shazeer2020glu}
Noam Shazeer. 2020.
\newblock {GLU} variants improve transformer.
\newblock \emph{arXiv preprint arXiv:2002.05202}.

\bibitem[{Vaswani et~al.(2017)Vaswani, Shazeer, Parmar, Uszkoreit, Jones,
  Gomez, Kaiser, and Polosukhin}]{vaswani2017attention}
Ashish Vaswani, Noam Shazeer, Niki Parmar, Jakob Uszkoreit, Llion Jones,
  Aidan~N Gomez, \L~ukasz Kaiser, and Illia Polosukhin. 2017.
\newblock \href
  {https://proceedings.neurips.cc/paper/2017/file/3f5ee243547dee91fbd053c1c4a845aa-Paper.pdf}
  {Attention is all you need}.
\newblock In \emph{Advances in Neural Information Processing Systems},
  volume~30, pages 5998--6008.

\bibitem[{Wenzek et~al.(2020)Wenzek, Lachaux, Conneau, Chaudhary, Guzm{\'a}n,
  Joulin, and Grave}]{wenzek2019ccnet}
Guillaume Wenzek, Marie-Anne Lachaux, Alexis Conneau, Vishrav Chaudhary,
  Francisco Guzm{\'a}n, Armand Joulin, and Edouard Grave. 2020.
\newblock \href {https://www.aclweb.org/anthology/2020.lrec-1.494} {{CCN}et:
  Extracting high quality monolingual datasets from web crawl data}.
\newblock In \emph{Proceedings of the 12th Language Resources and Evaluation
  Conference}, pages 4003--4012, Marseille, France. European Language Resources
  Association.

\bibitem[{Yang et~al.(2019)Yang, Zhang, Tar, and Baldridge}]{yang2019paws}
Yinfei Yang, Yuan Zhang, Chris Tar, and Jason Baldridge. 2019.
\newblock \href {https://doi.org/10.18653/v1/D19-1382} {{PAWS}-{X}: A
  cross-lingual adversarial dataset for paraphrase identification}.
\newblock In \emph{Proceedings of the 2019 Conference on Empirical Methods in
  Natural Language Processing and the 9th International Joint Conference on
  Natural Language Processing (EMNLP-IJCNLP)}, pages 3687--3692, Hong Kong,
  China. Association for Computational Linguistics.

\end{thebibliography}
\bibliographystyle{acl_natbib}

\begin{table*}[h!]
\begin{center}
\footnotesize
\begin{tabular}[b]{clrrr|clrrr}
\toprule
\textbf{ISO} & & \textbf{Tokens} & \textbf{Pages} & \textbf{mT5} & \textbf{ISO} & & \textbf{Tokens} & \textbf{Pages} & \textbf{mT5} \\
\textbf{Code} & \textbf{Language} & (B) & (M) & (\%) & \textbf{Code} & \textbf{Language} & (B) & (M) & (\%) \\
\cmidrule(r){1-5}\cmidrule(l){6-10}

en & English & 2,733 & 3,067 & 5.67 & mk & Macedonian & 1.8 & 2.1 & 0.62 \\
ru & Russian & 713 & 756 & 3.71 & ml & Malayalam & 1.8 & 2.1 & 0.62 \\
es & Spanish & 433 & 416 & 3.09 & mn & Mongolian & 2.7 & 2.1 & 0.62 \\
de & German & 347 & 397 & 3.05 & ur & Urdu & 2.4 & 1.9 & 0.61 \\
fr & French & 318 & 333 & 2.89 & be & Belarusian & 2.0 & 1.7 & 0.59 \\
it & Italian & 162 & 186 & 2.43 & la & Latin & 1.3 & 1.7 & 0.58 \\
pt & Portuguese & 146 & 169 & 2.36 & eu & Basque & 1.4 & 1.6 & 0.57 \\
pl & Polish & 130 & 126 & 2.15 & tg & Tajik & 1.4 & 1.3 & 0.54 \\
nl & Dutch & 73 & 96 & 1.98 & te & Telugu & 1.3 & 1.2 & 0.52 \\
tr & Turkish & 71 & 88 & 1.93 & fy & West Frisian & 0.4 & 1.1 & 0.51 \\
ja & Japanese & 164 & 87 & 1.92 & kn & Kannada & 1.1 & 1.1 & 0.51 \\
vi & Vietnamese & 116 & 79 & 1.87 & ky & Kyrgyz & 1.0 & 1.0 & 0.50 \\
id & Indonesian & 69 & 70 & 1.80 & sw & Swahili & 1.0 & 1.0 & 0.50 \\
cs & Czech & 63 & 60 & 1.72 & so & Somali & 1.4 & 0.9 & 0.48 \\
zh & Chinese & 39 & 55 & 1.67 & my & Burmese & 0.9 & 0.8 & 0.47 \\
fa & Persian & 52 & 54 & 1.67 & uz & Uzbek & 0.9 & 0.8 & 0.46 \\
ar & Arabic & 57 & 53 & 1.66 & km & Khmer & 0.6 & 0.8 & 0.46 \\
sv & Swedish & 45 & 49 & 1.61 & - & Russian (Latin) & 0.9 & 0.7 & 0.46 \\
ro & Romanian & 52 & 46 & 1.58 & sd & Sindhi & 1.6 & 0.7 & 0.45 \\
el & Greek & 43 & 42 & 1.54 & gu & Gujarati & 0.8 & 0.6 & 0.43 \\
uk & Ukrainian & 41 & 39 & 1.51 & - & Hindi (Latin) & 0.6 & 0.6 & 0.43 \\
hu & Hungarian & 39 & 37 & 1.48 & jv & Javanese & 0.3 & 0.6 & 0.42 \\
da & Danish & 29 & 29 & 1.38 & zu & Zulu & 0.2 & 0.6 & 0.42 \\
fi & Finnish & 25 & 27 & 1.35 & si & Sinhala & 0.8 & 0.5 & 0.41 \\
no & Norwegian & 27 & 25 & 1.33 & - & Japanese (Latin) & 0.3 & 0.5 & 0.41 \\
bg & Bulgarian & 22 & 23 & 1.29 & eo & Esperanto & 0.7 & 0.5 & 0.40 \\
hi & Hindi & 24 & 19 & 1.21 & co & Corsican & 0.2 & 0.5 & 0.40 \\
sk & Slovak & 18 & 18 & 1.19 & ga & Irish & 0.5 & 0.5 & 0.40 \\
ko & Korean & 26 & 16 & 1.14 & - & Greek (Latin) & 0.4 & 0.4 & 0.39 \\
th & Thai & 11 & 15 & 1.14 & - & Chinese (Latin) & 0.2 & 0.4 & 0.37 \\
ca & Catalan & 13 & 14 & 1.12 & pa & Punjabi & 0.6 & 0.4 & 0.37 \\
ms & Malay & 13 & 13 & 1.09 & ceb & Cebuano & 0.2 & 0.4 & 0.36 \\
iw & Hebrew & 17 & 12 & 1.06 & mg & Malagasy & 0.2 & 0.3 & 0.36 \\
lt & Lithuanian & 11 & 11 & 1.04 & ps & Pashto & 0.4 & 0.3 & 0.36 \\
sl & Slovenian & 8.8 & 8.5 & 0.95 & sn & Shona & 0.2 & 0.3 & 0.35 \\
mr & Marathi & 14 & 7.8 & 0.93 & gd & Scottish Gaelic & 0.4 & 0.3 & 0.35 \\
bn & Bengali & 7.3 & 7.4 & 0.91 & ku & Kurdish & 0.4 & 0.3 & 0.34 \\
et & Estonian & 6.9 & 6.9 & 0.89 & hmn & Hmong & 0.2 & 0.3 & 0.34 \\
lv & Latvian & 7.0 & 6.4 & 0.87 & su & Sundanese & 0.1 & 0.3 & 0.34 \\
az & Azerbaijani & 4.4 & 5.3 & 0.82 & ht & Haitian Creole & 0.2 & 0.3 & 0.33 \\
gl & Galician & 2.4 & 4.6 & 0.79 & ha & Hausa & 0.2 & 0.2 & 0.33 \\
cy & Welsh & 4.9 & 4.1 & 0.76 & ny & Chichewa & 0.1 & 0.2 & 0.29 \\
sq & Albanian & 4.0 & 4.1 & 0.76 & am & Amharic & 0.3 & 0.2 & 0.29 \\
ta & Tamil & 3.4 & 3.5 & 0.73 & - & Bulgarian (Latin) & 0.09 & 0.2 & 0.29 \\
sr & Serbian & 4.3 & 3.4 & 0.72 & yi & Yiddish & 0.3 & 0.1 & 0.28 \\
ne & Nepali & 3.2 & 2.9 & 0.69 & lo & Lao & 0.1 & 0.1 & 0.28 \\
lb & Luxembourgish & 1.0 & 2.7 & 0.68 & mi & Maori & 0.1 & 0.1 & 0.25 \\
hy & Armenian & 2.4 & 2.4 & 0.65 & sm & Samoan & 0.09 & 0.1 & 0.25 \\
kk & Kazakh & 3.1 & 2.4 & 0.65 & ig & Igbo & 0.09 & 0.09 & 0.24 \\
ka & Georgian & 2.5 & 2.3 & 0.64 & haw & Hawaiian & 0.09 & 0.08 & 0.24 \\
mt & Maltese & 5.2 & 2.3 & 0.64 & xh & Xhosa & 0.06 & 0.07 & 0.22 \\
af & Afrikaans & 1.7 & 2.2 & 0.63 & st & Sotho & 0.08 & 0.07 & 0.22 \\
fil & Filipino & 2.1 & 2.1 & 0.62 & yo & Yoruba & 0.05 & 0.05 & 0.20 \\
is & Icelandic & 2.6 & 2.1 & 0.62 \\
\bottomrule
\end{tabular}
\caption{Statistics of the mC4 corpus, totaling 6.6B pages and 6.3T tokens. The ``mT5'' column indicates the percentage of mT5 training data coming from a given language, using the default exponential smoothing value of $\alpha$=0.3. We list 107 ``languages'' as detected by \texttt{cld3}, but note six of these (marked ``Latin'') are just Romanized variants of existing languages.}
\label{tbl:mc4_languages}
\end{center}
\end{table*}

\clearpage

\foreach \x in {1,2,3,4,5}
 {\includepdf[page=\x,fitpaper]{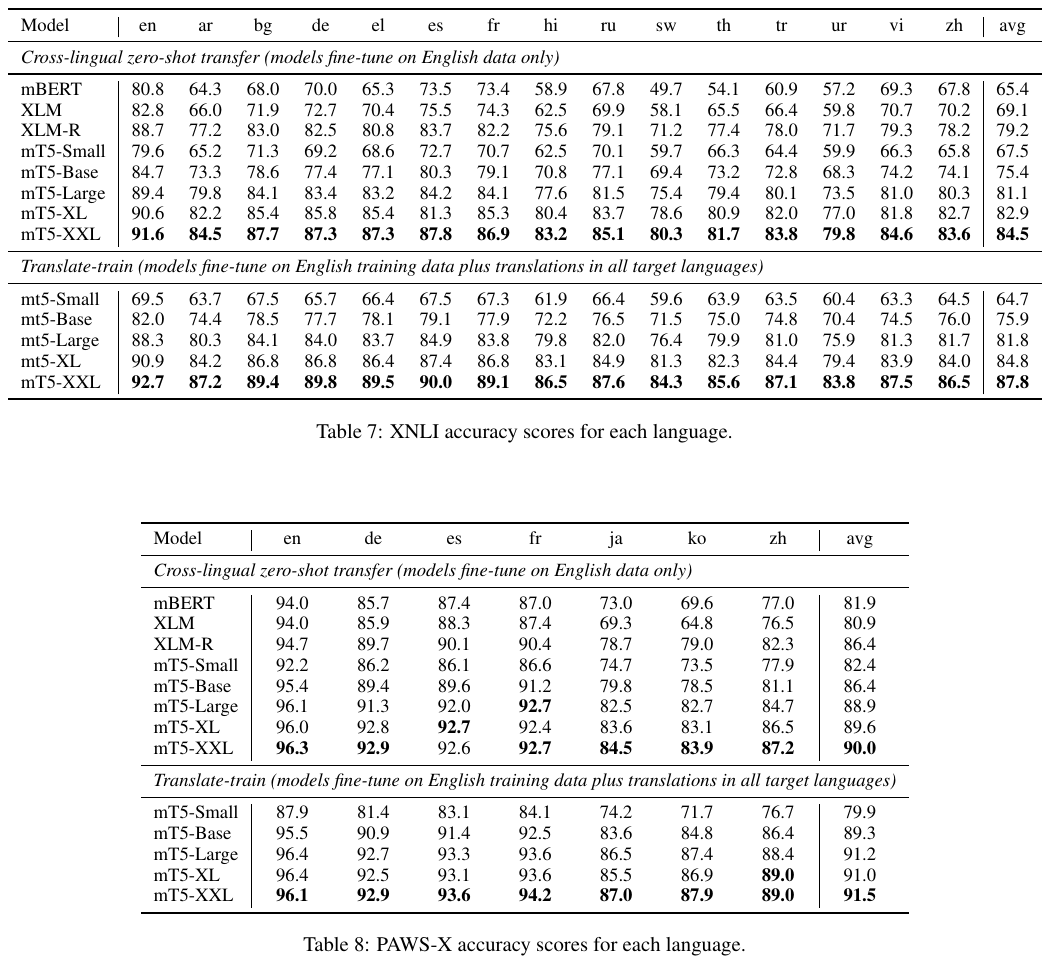}}


\end{document}